\documentclass[letterpaper, 10 pt, journal, twoside]{IEEEtran}  % Comment this line out if you need a4paper

\IEEEoverridecommandlockouts                              % This command is only needed if 
\usepackage{graphics} % for pdf, bitmapped graphics files
\usepackage{graphicx}
\usepackage{multirow}
\usepackage{cite}
\usepackage{amsmath} % assumes amsmath package installed
\usepackage{amssymb}  % assumes amsmath package installed
\usepackage{tikz}
\usepackage{lipsum}
\usepackage{siunitx}
\usepackage{hyperref}

\usepackage{amsthm}

\newtheorem{proposition}{Proposition}
\newtheorem{remark}{Remark}

\title{\LARGE \bf
Sensor Query Schedule and Sensor Noise Covariances for Accuracy-constrained Trajectory Estimation
}

\author{Abhishek Goudar$^{1}$ and Angela P. Schoellig$^{1}$% <-this % stops a space
\thanks{This work was supported in part by the Natural Sciences and Engineering Research Council of Canada (NSERC) and in part by the Canada CIFAR AI Chairs Program.}
\thanks{$^{1}$The authors are with the Learning Systems and Robotics Lab
 at the Technical University of Munich, Germany, and the University of Toronto Institute for Aerospace Studies, Canada. They are also affiliated with the University of Toronto Robotics Institute, the Munich Institute of Robotics and Machine Intelligence (MIRMI), and the Vector Institute for Artificial Intelligence. E-mail: abhishek.goudar@robotics.utias.utoronto.ca, angela.schoellig@tum.de}
\thanks{Digital Object Identifier (DOI): see top of this page.}
}

\newcommand{\Jpred}{\,\check{\rule{-0.6ex}{0ex}\mkern0mu \boldsymbol{J}}}

\begin{document}

\maketitle
%\thispagestyle{empty}
%\pagestyle{empty}

%%%%%%%%%%%%%%%%%%%%%%%%%%%%%%%%%%%%%%%%%%%%%%%%%%%%%%%%%%%%%%%%%%%%%%%%%%%%%%%%
\begin{abstract}
Trajectory estimation involves determining the trajectory of a mobile robot by combining prior knowledge about its dynamic model with noisy observations of its state obtained using sensors. The accuracy of such a procedure is dictated by the system model fidelity and the sensor parameters, such as the accuracy of the sensor (as represented by its noise covariance) and the rate at which it can generate observations, referred to as the sensor query schedule. Intuitively, high-rate measurements from accurate sensors lead to accurate trajectory estimation. However, cost and resource constraints limit the sensor accuracy and its measurement rate. Our work's novel contribution is the estimation of sensor schedules and sensor covariances necessary to achieve a \textit{specific estimation accuracy}. Concretely, we focus on estimating: (\textit{i}) the rate or \textit{schedule} with which a sensor of known covariance must generate measurements to achieve specific estimation accuracy, and alternatively, (\textit{ii}) the sensor covariance necessary to achieve specific estimation accuracy for a given sensor update rate. We formulate the problem of estimating these sensor parameters as semidefinite programs, which can be solved by off-the-shelf solvers. We validate our approach in simulation and real experiments by showing that the sensor schedules and the sensor covariances calculated using our proposed method achieve the desired trajectory estimation accuracy. Our method also identifies scenarios where certain estimation accuracy is unachievable with the given system and sensor characteristics.
\end{abstract}


\section{INTRODUCTION}

\begin{figure}[t]
	\centering

	\hspace*{-1.4em}\input{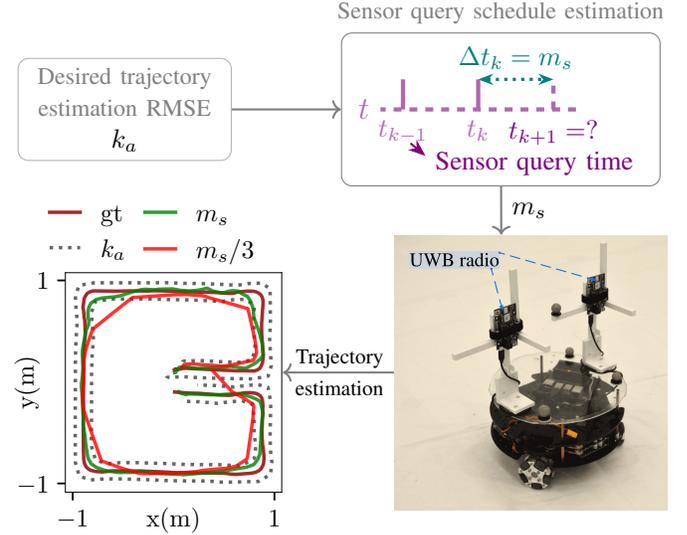}

	\vspace*{-1em}
	\caption{Application of the proposed sensor query schedule calculation approach to trajectory estimation. For a desired trajectory estimation root mean square error (RMSE), we calculate a sensor query schedule using our proposed method (top right). The calculated query schedule is then used to generate measurements on our robot for use in the trajectory estimation pipeline (bottom right). Results from real experiments showing the estimated trajectory, using the proposed query rate $m_s$\,(optimal rate) and a lower query rate $m_s/3$\,(suboptimal rate), the ground-truth (gt) trajectory, and the desired accuracy envelope ($k_a$) are shown (bottom left). The trajectory estimated with the proposed query rate is consistently within the accuracy envelope whereas the trajectory estimated using the lower query rate breaches the accuracy envelope on several instances, showing the validity of our approach.}
	\label{fig:method_overview}
	\vspace*{-1.em}
\end{figure}

\IEEEPARstart{T}{he} task of estimating the trajectory of a mobile system involves determining its position (and orientation) over time as it moves through the environment. A common approach to do this is to combine prior knowledge about the system's dynamics with measurements of the system's state obtained using sensors. The accuracy of any such trajectory estimation method is governed by \textit{(i)} the system model fidelity, and \textit{(ii)} the accuracy and frequency of sensory measurements. The accuracy of an unbiased sensor is captured by its \textit{sensor noise covariance} values, with lower covariance values implying higher accuracy. We refer to the frequency or schedule with which a sensor generates measurements as its \textit{sensor query schedule}. It follows that querying sensors with lower covariance at higher frequencies will achieve better trajectory estimation accuracy. Various factors such as manufacturing cost, power consumption, communication bandwidth, and available compute restrict the accuracy (or covariance) of a sensor and the rate at which it can generate measurements. For example, we consider ultra-wideband (UWB) radios that operate in two-way ranging (TWR) mode~\cite{dwm1000twr}. In this mode, a robot equipped with a UWB radio (referred to as a \textit{tag}) individually measures its distance to other UWB radios installed in the environment (referred to as \textit{anchors}). Due to the high-bandwidth nature of UWB technology, only one UWB radio can transmit at any given time. Additionally, due to the finite time of flight delay~\cite{dwm1000twr}, two UWB radios can only communicate up to a certain maximum measurement frequency. Hence, if a tag operates at its maximum measurement frequency, no other tag can acquire distance measurements. However, if the desired estimation accuracy can be achieved at a lower measurement frequency, then multiple tags can share the same channel. The overarching goal and contribution of our work is to estimate sensor parameters such as the sensor query schedule and the sensor noise covariances necessary to achieve a desired trajectory estimation accuracy.% enabling judicious use of sensors.

An overview of our approach to sensor query schedule estimation is shown in Figure~\ref{fig:method_overview}. For a desired trajectory estimation root mean square error (RMSE), we calculate the sensor query schedule using our proposed method (outlined in Section~\ref{sec:opt_prob}). The calculated query schedule is then used to generate measurements for trajectory estimation.

% Additionally, many real applications require a specific trajectory estimation accuracy that can be achieved with a less expensive setup. For example, consider an autonomous guided vehicle (AGV) operating in a warehouse that requires a maximum trajectory estimation root mean square error (RMSE) of 5\,\si{cm}. If a low-cost sensor with 10\,\si{cm} variance and 5\,\si{Hz} measurement frequency meets this requirement, it is perhaps preferable over an expensive sensor with 1\,\si{cm} variance and 2\,\si{Hz} measurement frequency that achieves similar accuracy. 
%Additionally, during operation, if a high-accuracy sensor is rendered ineffective, it may imperative to know if the same trajectory estimation accuracy can be achieved by increasing the measurement frequency of the remaining onboard sensors.
Next, we highlight another benefit of our approach. A common criterion for selecting sensor parameters is to choose parameters that minimize the state uncertainty~\cite{mourikisOptimalSensorScheduling2006,magnagoRobotLocalizationOdometryassisted2019}. However, such an approach cannot be used to determine if the desired estimation accuracy is achievable. Since our approach considers trajectory estimation accuracy explicitly, it can identify scenarios where certain estimation accuracy is infeasible given the system parameters. The following are the main contributions of our work:
\begin{itemize}
	\item An estimator-agnostic optimization-based approach for calculating sensor parameters that incorporates the desired estimation accuracy.
	\item Application of the proposed framework to estimate \textit{(i)} the sensor query schedule (for a given sensor noise covariance) and \textit{(ii)} sensor noise covariance (for a given query schedule) for trajectory estimation.
	\item Evaluation of our methods in simulation and real experiments. We share our code used in the experiments~\footnote{\url{https://github.com/abhigoudar/spi_acte}}.
\end{itemize}
\section{Related Work}

We review previous works that focus on sensor scheduling, sensor parameter estimation, and the achievable performance of estimators commonly used in state estimation for robotics. In probabilistic state estimation, a common approach for scheduling sensors in robotic systems is to optimize for sensor query frequencies that minimize the posterior state covariance estimator~\cite{mourikisOptimalSensorScheduling2006, yanMeasurementSchedulingCooperative2020}. Such an approach can yield low estimation error, but at the cost of injudicious use of sensory data. Our proposed method provides sensor query schedules optimized for a specific desired estimation accuracy, leading to judicious use of sensor measurements. An alternative approach to sensor scheduling is to trigger a measurement request when the estimated uncertainty of a particular estimator (the unscented Kalman filter (UKF)) is above a certain threshold~\cite{magnagoRobotLocalizationOdometryassisted2019}. Such an approach takes estimation uncertainty into account explicitly, but the calculated sensor schedule is specific to the employed state estimation method. In contrast, we use the posterior Cram{\'e}r-Rao bound (PCRB)~\cite{tichavskyPosteriorCramerRaoBounds1998} to calculate estimator-agnostic sensor schedules that achieve the desired estimation accuracy. The PCRB can be viewed as a Bayesian analog of the parametric Cram{\'e}r-Rao bound (CRB)~\cite{radhakrishnaraoInformationAccuracyAttainable1945,cramer1999mathematical}. Unlike the CRB, the PCRB also holds for estimators with unknown bias under mild assumptions~\cite{bergmanPosteriorCramerRaoBounds2001}. We also propose a \textit{convex} SDP formulation for calculating sensor query schedules that can be solved efficiently using off-the-shelf solvers and can identify scenarios where certain estimation accuracy is infeasible.

The Fisher information matrix (FIM)~\cite{Fisher_1925} and CRB~\cite{cramer1999mathematical,radhakrishnaraoInformationAccuracyAttainable1945} have been used in an array of problems such as the observability analysis of the simultaneous localization and mapping (SLAM) framework~\cite{wangObservabilityAnalysisSLAM2008}, planning for localization~\cite{papaliaPrioritizedPlanningCooperative2022}, and active perception~\cite{zhangPointCloudsFisher2019}. The CRB has been used to derive bounds on the achievable localization accuracy in range-based positioning~\cite{censiAchievableAccuracyRangefinder2007}. The CRB has also been applied to localization of transmitters from time-difference-of-arrival measurements~\cite{yangCramerRaoBoundOptimum2005,zhaoFindingRightPlace2022}, choosing among multiple observations by exploiting their \textit{submodular} (diminishing return property) nature~\cite{krauseNearoptimalObservationSelection2007}, and simultaneous placement and scheduling of sensors~\cite{krauseSimultaneousPlacementScheduling2009}. More recently, the problem of optimal beacon placement for range-aided localization was formulated as maximization of a submodular set function based on the FIM~\cite{kavetiOASISOptimalArrangements2024}. In a related application, an information-theoretic objective is used to design optimal placement for a set of diverse sensors on a mobile robot for improved sensing~\cite{sequeiraOptimalBeaconPlacement2024}. The PCRB has been relatively less explored in robotics. A common application of the PCRB is to design a generic framework to deploy and schedule multisensor systems based on sensor contributions~\cite{hernandezMultisensorResourceDeployment2004}. Since PCRB provides a lower bound on the estimation mean square error, it has been used to evaluate the effect of different sensing modalities for SLAM~\cite{selvaratnamEffectSensorModality2016}. The previous approaches directly minimize the lower bounds provided by the CRB and the PCRB to estimate sensor placements and schedules, and do not consider the required estimation accuracy explicitly. Our proposed method calculates an estimator-agnostic sensor query schedule by minimizing the predictive PCRB that incorporates the required estimation accuracy explicitly.

Sensor covariances are typically identified using sensor data sheets or through an offline calibration procedure. An alternative approach is to estimate the sensor noise models online using a bi-level optimization approach. Specifically, covariances for robot motion and sensor models can be estimated using an expectation-maximization-based approach~\cite{wongVariationalInferenceParameter2020}. Alternatively, a data-driven approach can be used to estimate the process density of a motion prior using ground-truth information~\cite{wongDataDrivenMotionPrior2020}. Similarly, sensor noise models can also be estimated using incremental maximum a posteriori inference~\cite{  qadriLearningCovariancesEstimation2024}. More recently,~\cite{khosoussiJointStateNoise2025} proposed a convex formulation for the problem of joint state and noise covariance estimation within the maximum a posteriori (MAP) inference framework. While these works proposed methods that estimate sensor covariances to improve trajectory estimation accuracy, our framework estimates sensor noise covariances required to achieve a specific estimation accuracy. As mentioned before, our proposed framework can also be used to identify if particular estimation accuracy is unachievable for a given sensor setup. To the best of the authors' knowledge, the calculation of sensor query schedules and sensor noise covariances that achieve a desired trajectory estimation accuracy has not been done before. In the next section we introduce the necessary background for our problem formulation.

%\textcolor{blue}{
%TODO: The outline for the next sections is as follows. We provide the relevant background theory in Section~\ref{sec:preliminaries} and formulate our problems the system process and measurement model in Section~\ref{sec:sys_model} and the estimation error metric in Section~\ref{sec:mse_lb}. Next, we present a recursive lower bound to estimation error in Section~\ref{sec:rec_pcrb}. In Section~\ref{sec:acc_const}, we incorporate the desired estimation accuracy into the lower bound, and formulate our main optimization problem in Section~\ref{sec:opt_prob}. 
%}

%
\section{Preliminaries}

We introduce the notation used in this paper. Lower-case symbols are used to represent scalar quantities. Bold lower-case and upper-case symbols are used to represent vectors and matrices, respectively. The set of positive real numbers is denoted by $\mathbb{R}_{++}$ and the sets of $d \times d$ positive definite and positive semidefinite matrices are represented by $\mathbb{S}_{++}^d$ and $\mathbb{S}^d_+$, respectively. For a matrix $\mathbf{A} \in \mathbb{R}^{d \times d}$,  $\mathbf{A} \succcurlyeq 0$ implies $\mathbf{A} \in \mathbb{S}^d_{+}$, and $\mathbf{A} \succ 0$ implies $\mathbf{A} \in \mathbb{S}^d_{++}$. For matrices $\mathbf{A}, \mathbf{B} \in \mathbb{S}^d_{+}$, $\mathbf{A} \succcurlyeq \mathbf{B}$ implies $\mathbf{A} - \mathbf{B} \succcurlyeq 0$. 

\subsection{System model} \label{sec:sys_model}

We consider nonlinear systems of the form
\begin{align}
    \begin{split}
        \mathbf{x}_{k+1} &= f(\mathbf{x}_{k}, \mathbf{u}_k) + \mathbf{w}_k, \\
        \mathbf{y}_k &= h(\mathbf{x}_k) + \boldsymbol{\eta}_k,
        \label{eqn:system_model}            
    \end{split}
\end{align}
where the subscript $k$ represents the time index, $\mathbf{x}_k$ is state at time $t_k$, $\mathbf{u}_k$ is the control input, $\mathbf{y}_k$ is the measurement obtained at time $t_k$, and $\mathbf{w}_k \sim \mathcal{N}(\mathbf{0}, \mathbf{Q}_k)$ and $\boldsymbol{\eta}_k \sim \mathcal{N}(0, \mathbf{R}_k)$ are additive white Gaussian noise (AWGN) with covariances $\mathbf{Q}_k$ and $\mathbf{R}_k$, respectively. The system process model is represented by the nonlinear function $f(\cdot)$ and the measurement model by $h(\cdot)$, respectively. We assume that the measurement noise, process noise, and the initial state, $\mathbf{x}_0$, are uncorrelated with each other and over time.

\subsection{Estimation error} \label{sec:mse_lb}

The goal of a state estimator is to infer the state of a system by combining knowledge about the system model with noisy observations of the state obtained using sensors. To assess its performance, we can use the estimator \textit{mean squared error} (MSE) as the metric. Let $g(\mathbf{y})$ be an estimate of the true state $\mathbf{x}$, then the estimator MSE is given by

\begin{equation}
    \mathrm{MSE}(\mathbf{x}) = \mathbb{E}\left[ \left( g(\mathbf{y}) - \mathbf{x} \right)  \left( g(\mathbf{y}) - \mathbf{x} \right)^T \right], \label{eqn:mse}
\end{equation}
which is matrix-valued and considers both the (co)variance and the bias of an estimator~\cite[Section 8.2]{wackerlyMathematicalStatisticsApplications2008}. An estimate of the state can be obtained using a filtering or an optimization-based approach. We will refer to the matrix resulting from the expected outer product in~\eqref{eqn:mse} as the \textit{error correlation} matrix. 

\subsection{Posterior Cram\'er-Rao bound}\label{sec:pcrb_intro}

The PCRB~\cite{van2004detection} provides an estimator-agnostic lower bound to the estimation error~\eqref{eqn:mse}. More importantly for us, PCRB sets a lower limit on the estimator MSE in the sense that the difference between the expected mean square error correlation matrix and a matrix determined by the model of the estimation problem is positive semidefinite~\cite{bergmanPosteriorCramerRaoBounds2001}. Consider the sequence of states and measurements until time $k$: $\mathbf{x}_{0:k} = \left( \mathbf{x}_0, \mathbf{x}_1,...,\mathbf{x}_k \right)$, $\mathbf{y}_{1:k} = \left( \mathbf{y}_1,...,\mathbf{y}_k \right)$. The joint distribution over the state and measurements based on \eqref{eqn:system_model} is
\begin{equation*}
    p(\mathbf{x}_{0:k}, \mathbf{y}_{1:k} | \theta) = p(\mathbf{x}_0) \prod_{i=0}^{k-1} p(\mathbf{x}_{i+1} | \mathbf{x}_{i}, \theta ) \prod_{j=1}^k p(\mathbf{y}_j | \mathbf{x}_j, \theta) 
    %\label{eqn:batch_pdf}
\end{equation*}
where $\theta$ represents sensor parameters that we want to identify. 
%The conditional distributions are derived from the system model~\eqref{eqn:system_model}:
% \begin{align*}
%     p(\mathbf{x}_{i+1} | \mathbf{x}_{i}, \theta ) \propto \exp \frac{1}{2} \left( \mathbf{x}_{i+1} - f(\mathbf{x}_i, \mathbf{u}_i) \right)^T \mathbf{Q}_i^{-1} \left( \mathbf{x}_{i+1} - f(\mathbf{x}_i, \mathbf{u}_i) \right)
% \end{align*}
%
The PCRB on the estimation error is
\begin{align*}
    \boldsymbol{P}_{\mathbf{x}_{0:k}}(\theta) &\triangleq \mathbb{E} \left[  \left( g(\mathbf{y}_{1:k}) - \mathbf{x}_{0:k} \right)  \left( g(\mathbf{y}_{1:k}) - \mathbf{x}_{0:k} \right)^T  \right] \\ &\succcurlyeq  ({\mathcal{I}_{\mathbf{x}_{0:k}}(\theta)})^{-1}, 
    \label{eqn:pcrb}
\end{align*}
where $\boldsymbol{P}_{\mathbf{x}_{0:k}}$ is the error correlation matrix associated with estimating $\mathbf{x}_{0:k}$ and
\begin{equation}
    \mathcal{I}_{\mathbf{x}_{0:k}}(\theta) = \mathbb{E} \left[ - \frac{\partial^2 \log p(\mathbf{x}_{0:k}, \mathbf{y}_{1:k} | \theta)}{\partial \mathbf{x}_{0:k} \partial \mathbf{x}_{0:k}^T} \right]
    \label{eqn:fim}
\end{equation}
is the (Fisher) information matrix~\cite{Fisher_1925,radhakrishnaraoInformationAccuracyAttainable1945}. Note that the information matrix is a function of the parameters $\theta$. 

\section{Problem statement} \label{sec:prob_statement}

We represent a sensor query schedule either by a sequence of monotonic time instants, $\boldsymbol{t}_s = \left(t_1, t_2,...\right)$, or as a single parameter, $m_s$, when the sensor is queried at a constant rate. The sensor covariance is represented by $\mathbf{R}$ and the desired estimation accuracy by the positive scalar $k_a$. 

The objectives of our works are \textit{(i)} identify a sensor query schedule (for a given sensor covariance) with minimum entries in $\boldsymbol{t}_s$ or the smallest $m_s$ that achieves the desired estimation accuracy, and \textit{(ii)} identify sensor covariances (for a given sensor query schedule) that achieve the desired estimation accuracy. We can express these objectives as the following optimization problem
\begin{equation}
	\begin{split}
		\underset{\theta}{\min} & \quad  s(\theta) \\
		\mathrm{s.t.} & \quad \mathrm{MSE}(\mathbf{x}_{0:k}) ~ \mathrm{is ~  at ~ most ~} k_a^2.
	\end{split}
	\label{eqn:generic_opt_problem}
\end{equation}
where as before $s(\cdot)$ is a parameter-dependent function with $\theta = m_s$ or $\theta = \mathbf{R}$. In the next section, we develop a computationally tractable convex formulation for~\eqref{eqn:generic_opt_problem}.

%As mentioned previously, the objectives of our work are \textit{(i)} given a sensor with known covariance, identify a query schedule with minimum queries to achieve a pre-specified estimation accuracy, and \textit{(ii)} for a given sensor query schedule, estimate the sensor covariances necessary to achieve a pre-specified estimation accuracy.

%A sensor query schedule is represented either by a sequence of nondecreasing monotonic time instants, $\boldsymbol{t}_s = \left(t_1, t_2,...\right)$, or as a single frequency parameter, $m_s$, where the sensor is queried at a constant rate. Our objective is to identify a query schedule with minimum entries in $\boldsymbol{t}_s$ or the smallest possible $m_s$ that achieves the required estimation accuracy. For example, this could be the desired localization root-mean-squared error. We assume that the system motion model, sensor measurement model, and their corresponding Jacobians can be calculated analytically. 	

%
\section{Methodology}\label{sec:methodology}

A challenge associated~\eqref{eqn:generic_opt_problem} is that the MSE depends on a particular realization $g(\mathbf{y}_{1:k})$ of a specific estimator. In contrast, the FIM provides an estimator-independent lower bound to the estimator MSE. An alternative optimization problem that uses the FIM to estimate sensor parameters is as follows
\begin{equation}
	\begin{split}
		\underset{\theta}{\min} & \quad  s(\theta) \\
		\mathrm{s.t.} & \quad {({\mathcal{I}_{\mathbf{x}_{0:k}}(\theta)})^{-1}} ~ \mathrm{is ~  at ~ most ~} k_a^2,
	\end{split}
	\label{eqn:naiv_inf_opt}
\end{equation}
However, calculating the inverse of the FIM can be expensive. In the next section, we derive a recursive form for the PCRB that is computationally tractable and in the subsequent section we incorporate the desired estimation accuracy into the recursive PCRB.
\subsection{Recursive posterior Cram\'er-Rao bound} \label{sec:rec_pcrb}

Our objective in this section is to obtain a recursive form for the PCRB associated with the joint distribution $p(\mathbf{x}_{0:k}, \mathbf{y}_{1:k}, \mathbf{x}_{k+1} | \theta)$. For brevity, we denote the joint distribution $p(\mathbf{x}_{0:k}, \mathbf{y}_{1:k}, \mathbf{x}_{k+1} | \theta)$ by $p_{k+1}$. Let $\check{\mathcal{I}}_{\mathbf{x}_{0:k+1}}(\theta)$ be the information matrix derived from $p_{k+1}$. The \textit{check} $\left( \, \check \, \, \right)$ notation is used to indicate the information matrix associated with a predictive distribution, i.e., the information until time $k=n+1$ excluding the observation $\mathbf{y}_{k+1}$. The PCRB for a decomposed form of the error correlation matrix and the information matrix is
\begin{align*}
	\begin{bmatrix}
		\check{\boldsymbol{P}} _{\mathbf{x}_{0:k}} & \check{\boldsymbol{P}} _{\mathbf{x}_{0:k}, \mathbf{x}_{k+1}} \\
		\check{\boldsymbol{P}}_{\mathbf{x}_{k+1}, \mathbf{x}_{0:k}}  & \check{\boldsymbol{P}}_{\mathbf{x}_{k+1}} 
	\end{bmatrix} \succcurlyeq
	\begin{bmatrix}
		\boldsymbol{A}_{k+1}  & \boldsymbol{B}_{k+1}  \\
		\boldsymbol{B}_{k+1}^T & \boldsymbol{C}_{k+1}
	\end{bmatrix}^{-1},
\end{align*}
where the dependency on $\theta$ has been omitted for brevity, $\check{\boldsymbol{P}}_{\mathbf{x}_{0:k}}$ and $\check{\boldsymbol{P}}_{\mathbf{x}_{k+1}}$ are the error correlation submatrices associated with $\mathbf{x}_{0:k}$ and $\mathbf{x}_{k+1}$, respectively, $\check{\boldsymbol{P}}_{\mathbf{x}_{0:k}, \mathbf{x}_{k+1}}$ is the error cross-correlation submatrix, and
\scalebox{0.995}{\parbox{.5\linewidth}{%
\begin{align*}
	\boldsymbol{A}_{k+1} &= \mathbb{E} \left[-\frac{\partial^2 \log p_{k+1}}{\partial \mathbf{x}_{0:k} \partial \mathbf{x}_{0:k}^T}\right],
	\boldsymbol{B}_{k+1}  = \mathbb{E} \left[-\frac{\partial^2 \log p_{k+1}}{\partial \mathbf{x}_{0:k} \partial \mathbf{x}_{k+1}^T}\right],\\
	\boldsymbol{B}_{k+1}^T  &= \mathbb{E} \left[-\frac{\partial^2 \log p_{k+1}}{\partial \mathbf{x}_{k+1} \partial \mathbf{x}_{0:k}^T}\right],
	\boldsymbol{C}_{k+1}  = \mathbb{E} \left[-\frac{\partial^2 \log p_{k+1}}{\partial \mathbf{x}_{k+1} \partial \mathbf{x}_{k+1}^T}\right].
\end{align*}
}}
As noted in~\cite{tichavskyPosteriorCramerRaoBounds1998}, the estimation error associated with $\mathbf{x}_{k+1}$ is lower bounded by the inverse of the right-lower block of $(\check{\mathcal{I}}_{\mathbf{x}_{0:k+1}}(\theta))^{-1}$, denoted here by $\Jpred_{\mathbf{x}_{k+1}}$:
\begin{align}
	\check{\boldsymbol{P}}_{\mathbf{x}_{k+1}} \succcurlyeq (\Jpred_{\mathbf{x}_{k+1}})^{-1} \triangleq  \boldsymbol{C}_{k+1} - \boldsymbol{B}_{k+1}^T \boldsymbol{A}_{k+1}^{-1} \boldsymbol{B}_{k+1}.
	\label{eqn:np1_lb_batch}
\end{align}
However, \eqref{eqn:np1_lb_batch} requires inverting a relatively large $\boldsymbol{A}_{k+1}$, which is expensive. The following proposition gives a recursive form that is computationally more efficient.

\begin{proposition}
	Let $\check{\mathbf{x}}_{k+1}$ be the one-step-ahead predictor for $\mathbf{x}_{k+1}$. The correlation matrix of estimation error $\check{\boldsymbol{P}}_{\mathbf{x}_{k+1}} = \mathbb{E} \left[ (\check{\mathbf{x}}_{k+1} - \mathbf{x}_{k+1})(\check{\mathbf{x}}_{k+1} - \mathbf{x}_{k+1})^T\right] $  is lower bounded by matrix $\Jpred_{\mathbf{x}_{k+1}}$ for any time $k=0,1,2,\dots$:
	\begin{align}
		\check{\boldsymbol{P}}_{\mathbf{x}_{k+1}} \succcurlyeq (\Jpred_{\mathbf{x}_{k+1}})^{-1},
		\label{eqn:pcrb_np1}
	\end{align}
	where $\Jpred_{\mathbf{x}_{k+1}}$ obeys the recursion
	\begin{equation}
		\Jpred_{\mathbf{x}_{k+1}} = \boldsymbol{D} ^{22}_{k+1} - \boldsymbol{D} ^{21}_{k+1} \left( \boldsymbol{D}^{11}_{k+1} + \Jpred_{\mathbf{x}_k} \right)^{-1}  \boldsymbol{D} ^{12}_{k+1}
		\label{eqn:recr_pcrb_eqn}
	\end{equation}
	with
	\begin{align*}
		\boldsymbol{D} ^{11}_{k+1} &= \mathbb{E} \left[ - \frac{\partial^2 \log p(\mathbf{x}_{k+1} | \mathbf{x}_k)}{\partial \mathbf{x}_k \partial \mathbf{x}_k^T } - \frac{\partial^2 \log p(\mathbf{y}_{k} | \mathbf{x}_k)}{\partial \mathbf{x}_k \partial \mathbf{x}_k^T } \right], \\
		\boldsymbol{D} ^{12}_{k+1} &= \mathbb{E} \left[ - \frac{\partial^2 \log p(\mathbf{x}_{k+1} | \mathbf{x}_k)}{\partial \mathbf{x}_k \partial \mathbf{x}_{k+1}^T } \right], 
		\quad \boldsymbol{D} ^{21}_{k+1} = \left( \boldsymbol{D} ^{12}_{k+1} \right)^T, \\
		\boldsymbol{D} ^{22}_{k+1} &= \mathbb{E} \left[ - \frac{\partial^2 \log p(\mathbf{x}_{k+1} | \mathbf{x}_k)}{\partial \mathbf{x}_{k+1} \partial \mathbf{x}_{k+1}^T } \right].
	\end{align*}
	The recursion is initiated by $\Jpred_{\mathbf{x}_0} = \mathbb{E} \left[- \frac{\partial^2 \log p(\mathbf{x}_{0})}{\partial \mathbf{x}_{0} \partial \mathbf{x}_{0}^T } \right]$. The statement holds under the additional assumption that the relevant derivatives, expectations, and matrix inversions exist. 
	\label{prop:recr_pcrb}
	\vspace*{-1em}
	\begin{proof}
		The result follows the same argument as~\cite[Proposition 1]{tichavskyPosteriorCramerRaoBounds1998} applied to the prediction step, instead of the update step, of the filter.
		% The procedure for the proof is similar to that of Proposition 1 in~\cite{tichavskyPosteriorCramerRaoBounds1998}, with the exception that the joint distribution in our case is the predictive distribution over $\mathbf{x}_{k+1}$, $p(\mathbf{X}_k , \mathbf{Y}_k , \mathbf{x}_{k+1} | \theta)$.
	\end{proof}    
\end{proposition}

Although the Hessians in~\eqref{eqn:recr_pcrb_eqn} look complicated, they simplify for the system~\eqref{eqn:system_model} under the AWGN assumption:
\begin{equation}
	\begin{split}
		{\boldsymbol{D} }^{11}_{k+1} &= \mathbb{E} \left[ \mathbf{F}_k^{T} \mathbf{Q}_{k}^{-1}  \mathbf{F}_k \right] + \mathbb{E} \left[ \mathbf{H}_k^T \mathbf{R}^{-1}  \mathbf{H}_k \right], \\ 
		{\boldsymbol{D} }^{12}_{k+1} &= - \mathbb{E} \left[ \mathbf{F}_k \right] \mathbf{Q}_{k}^{-1},
		\quad {\boldsymbol{D} }^{22}_{k+1} = \mathbf{Q}_{k}^{{-1}},
	\end{split}
	\label{eqn:pcrb_hessians}
\end{equation}
where
\begin{align*}
	\mathbf{F}_k = \frac{\partial f(\mathbf{x}_{k})}{\partial \mathbf{x}_{k}}, \quad \mathbf{H}_k = \frac{\partial h(\mathbf{x}_{k})}{\partial \mathbf{x}_{k}}.
\end{align*}
The expectations can be computed using Monte Carlo methods or Gaussian quadrature. The upshot is that we now have a way to recursively compute a lower bound to the estimation error at each time step.
%Additionally, upper or lower limits on required estimation accuracy can easily be obtained from system or task specifications.

\begin{figure}[t]
	\centering
	\scalebox{0.8}{
			    \begin{tikzpicture}[arrow/.style={red!80!black,ultra thick,->,>=latex}]
        % accuracy bound
        \draw[color=black!60, fill=blue!10, thick](0,0) circle (1.6cm) node[](acc_bound){};
        % error ellipse
        \draw[rotate=60, color=black!60, fill=green!20, solid, very thick] (0,0) ellipse [x radius=1.59cm, y radius=0.7cm] node[](err_elps){};
        \draw [color=black,thin] (0,0) -- (60:1.6cm);
        \node [color=black, anchor=south west, xshift=0.16cm, yshift=0.15cm, rotate=60] {\large $\sqrt{\lambda}_{\mathrm{max}}$};
        \draw [color=black,thin] (0,0) -- (330:0.69cm);
        \node [color=black, anchor=north east, xshift=-0.03cm, yshift=0.13cm, rotate=60] {\large $\sqrt{\lambda}_{\mathrm{min}}$};
        \draw [color=black,fill=black] (0,0) circle (2pt) node[black, below=1mm]{};
		\draw [color=black,thin] (0,0) -- (17:1.59cm);
		\draw [color=black, anchor=west] (0.96, 0.11) node {\large $k_a$};
        \node[draw=orange,fill=orange!10,rounded corners,text width=2.5cm] (err_elps_exp) at (-3.5cm,-0.5cm) {Error correlation ellipse};
        \node[draw=orange,fill=orange!10,rounded corners,text width=1.5cm] (acc_bound_exp) at (3.5cm, 1cm) {Accuracy bound};
        \draw[arrow](err_elps_exp.east) to ([xshift=-5.4ex, yshift=-2ex]{err_elps}); 
        \draw[arrow](acc_bound_exp.west) to ([xshift=8ex, yshift=5ex]{acc_bound});
        \draw [gray, dashed, ultra thick] plot [smooth, tension=1] coordinates { (4, -1) (3,-0.5) (0,0) (-3, 0.5) (-4,0.8)} node [above]{\Large$t$}; 
	\end{tikzpicture}
	}
	\caption{The error correlation matrix, $\check{\boldsymbol{P}}_{\mathbf{x}_{t}}$, viewed as the ellipse $\{ \mathbf{x} \in \mathbb{R}^2 \mid \mathbf{x}^T (\check{\boldsymbol{P}}_{\mathbf{x}_{t}})^{-1} \mathbf{x} \leq 1\}$ (green), represents the spread of errors and the square root of its maximum eigenvalue, $\lambda_{\max}$, gives the maximum error along any principle axis. To achieve desired accuracy $k_a$, the error ellipse must be within the accuracy bound ellipse~$\{ \mathbf{x} \in \mathbb{R}^2 \mid \mathbf{x}^T ({k_a^2 \mathbf{I}})^{-1} \mathbf{x} \leq 1\}$ (blue) at each time step $t$ (gray dashed line).}
	\label{fig:err_acc_const}
	\vspace*{-1.5em}
\end{figure}
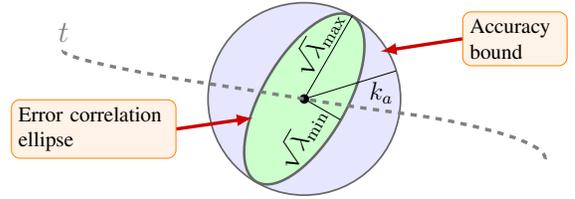

\subsection{Constraint on estimation error} \label{sec:acc_const}
In this section, we take the required estimation accuracy into account by incorporating it as a constraint on the information submatrix $\Jpred_{\mathbf{x}_{k+1}}$. We begin by noting that the square root of the eigenvalues of the error correlation matrix $\check{\boldsymbol{P}}_{\mathbf{x}_{k+1}}$ represent the magnitude of the spread of estimation error along principle directions (given by eigenvectors), with the square root of the maximum eigenvalue, $\sqrt{\lambda}_{\max}({\check{\boldsymbol{P}}}_{\mathbf{x}_{k+1}})$, representing the maximum spread of estimation error in any direction (see Figure~\ref{fig:err_acc_const}). To achieve the necessary accuracy, we require that the square root of the maximum eigenvalue of the error correlation matrix is at most the required accuracy:
\begin{equation}
	\lambda_{\max}({\check{\boldsymbol{P}}}_{\mathbf{x}_{k+1}}) \leq k_a^2.
	\label{eqn:eig_cond}
\end{equation}
The above inequality can also be expressed by the following convex constraint~\cite[Appendix A.5.2]{boydConvexOptimization2004}
\begin{equation}
	{\check{\boldsymbol{P}}}_{\mathbf{x}_{k+1}} \preccurlyeq k_a^2 \, \mathbf{I},
	\label{eqn:eig_mat_cond}
\end{equation}
where $\mathbf{I}$ is the identity matrix. We can obtain a constraint on the information submatrix from~\eqref{eqn:pcrb_np1} and~\eqref{eqn:eig_mat_cond} as
\begin{equation}
	\Jpred_{\mathbf{x}_{k+1}} \succcurlyeq (\check{\boldsymbol{P}}_{\mathbf{x}_{k+1}})^{-1} \succcurlyeq k_a^{-2} \mathbf{I}.
	\label{eqn:inf_err_acc_inq}
\end{equation}
Including the recursion from~\eqref{eqn:recr_pcrb_eqn} in the above inequality gives the following relation
\begin{align*}
	\boldsymbol{D} ^{22}_{k+1} - \boldsymbol{D} ^{21}_{k+1} \left( \boldsymbol{D}^{11}_{k+1} + \Jpred_{\mathbf{x}_k} \right)^{-1}  \boldsymbol{D} ^{12}_{k+1} \succcurlyeq k_a^{-2} \mathbf{I}.
\end{align*}
Moving $k_a^{-2} \mathbf{I}$ to the left side and using the Schur complement characterization of symmetric positive semidefinite matrices~\cite[Proposition 2.2]{gallierSchurComplementSymmetric2010}, we get the following conic inequality constraint
\begin{equation} 
	\boldsymbol{S}_{k}(\theta) \triangleq
	\begin{bmatrix}
		\Jpred_{\mathbf{x}_k} + \boldsymbol{D}^{11}_{k+1}  &  \boldsymbol{D}^{12}_{k+1} \\
		\boldsymbol{D}^{21}_{k+1} & \boldsymbol{D}^{22}_{k+1} - k_a^{-2} \mathbf{I} 
	\end{bmatrix} \succcurlyeq 0. 
	\label{eqn:inf_err_acc_mat_const}
\end{equation}

\subsection{Optimization problem} \label{sec:opt_prob}
We now present the main result of our work. As outlined in Section~\ref{sec:prob_statement}, our objective is to identify sensor parameters to achieve a pre-specified estimation accuracy. To achieve the necessary estimation accuracy, \eqref{eqn:inf_err_acc_mat_const} outlines the constraint that must be satisfied. Any parameter that satisfies~\eqref{eqn:inf_err_acc_mat_const} is a feasible parameter. As such, we can cast the problem of identifying optimal parameters at each time step $k=0,1,\dots$ as solving the following optimization problem
\begin{equation}
	\begin{split}
		\underset{\theta}{\min} & \quad  s(\theta) \\
		\mathrm{s.t.} & \quad \boldsymbol{S}_{k}(\theta) \succcurlyeq 0.
	\end{split}
	\label{eqn:param_opt_problem}
\end{equation}
where as before $s(\cdot)$ is a parameter-dependent function. The parameter $\theta$ can be constant or change over time. The optimization problem~\eqref{eqn:param_opt_problem} is general and can be used in any setup that satisfies the assumptions outlined in Section~\ref{sec:prob_statement}. In the next section, we
apply it to the calculation of sensor query schedules and sensor covariances for estimating the trajectory of a mobile robot to the desired accuracy.

%\begin{remark}
%		Note that the constraint in~\eqref{eqn:param_opt_problem} considers only the outer inequality $\Jpred_{\mathbf{x}_{k+1}} \succcurlyeq k_a^{-2} \mathbf{I}$ in~\eqref{eqn:inf_err_acc_inq}. As such, solving~\eqref{eqn:param_opt_problem} does not guarantee $ \check{\boldsymbol{P}}_{\mathbf{x}_{k+1}} \preceq k_a^{2} \mathbf{I}$. However, we observed in our experiments that in the case where ~\eqref{eqn:param_opt_problem} is feasible, we have $\Jpred_{\mathbf{x}_{k+1}} = k_a^{-2} \mathbf{I}$, and hence the inequalities in~\eqref{eqn:inf_err_acc_inq} are satisfied with equality.
%\end{remark}
%
\section{Trajectory Estimation}

Our approach in this section is as follows: First, we present expressions for the process model Jacobian, $\mathbf{F}_k$, the measurement model Jacobian, $\mathbf{H}_{k}$, and the covariance function, $\mathbf{Q}_k$, that are required for formulating the constraint~\eqref{eqn:inf_err_acc_mat_const}. Next, we use this constraint in~\eqref{eqn:param_opt_problem} to solve for sensor query schedules and sensor covariances. We use the framework of continuous-time trajectory estimation~\cite{barfootgp2014,barfootStateEstimationRobotics2024} that uses a continuous-time motion model and discrete-time measurement models. We use a continuous-time motion model as the state and the corresponding uncertainty can be computed at arbitrary time instants, which is ideal for scheduling measurements. 

\subsection{Motion model} \label{sec:wnov_mot_model}

We use a \textit{white noise on velocity} (WNOV) or a constant-velocity motion model as the system process model. Specifically, we use motion priors based on linear time-invariant stochastic differential equations: 
\begin{align}
    \dot{\mathbf{x}}(t) = \mathbf{u}(t) + \mathbf{w}(t),
    \label{eqn:wnov_proc_model}
\end{align}
where $\mathbf{x}(t) \in \mathbb{R}^d$ is the state and $\dot{\mathbf{x}}(t)$ is the corresponding time derivative, $\mathbf{u}(t) \in \mathbb{R}^{d}$ is the velocity input, and $\mathbf{w}(t) \sim \mathcal{GP}(0, \boldsymbol{Q} \delta(t - t') )$ is a zero-mean Gaussian Process (GP) with power spectral density matrix $\boldsymbol{Q}$. The state consists of the robot's position, $\mathbf{x}(t) = \mathbf{p}(t)$, however, the procedure outlined here can be applied to nonlinear spaces such as the special Euclidean ($SE(d)$) manifold using local
poses defined on vector spaces~\cite{Anderson2015}. The mean of the GP prior between two consecutive time instants $t_k$ and $t_{k+1}$ is
\begin{align}
    \mathbf{x}_{k+1} &= \mathbf{x}_{k} + \mathbf{u}_k \Delta t_{k},
    \label{eqn:wnov_gp_prior_mean}
\end{align}
where $\mathbf{x}_{k} = \mathbf{x}(t_k)$ and $\Delta t_{k} = t_{k+1} - t_k$. The process model Jacobian and the covariance function are~\cite[Section 3.4]{barfootStateEstimationRobotics2024}:
\begin{align}
    \mathbf{F}_k &= \mathbf{I}, \quad \mathbf{Q}_{k} = \boldsymbol{Q} \Delta t_{k}.
    \label{eqn:wnov_jac_cov}
\end{align}
\subsection{Measurement models}
In this section, we present measurement models and Jacobians for two commonly used sensing modalities: \textit{position} and \textit{range} sensors.  

\subsubsection{Position measurement model} \label{sec:pos_meas_model}
The observations in this case are the position of the mobile robot over time:
\begin{align}
    \tilde{\mathbf{p}}_k = \mathbf{x}_k + \boldsymbol{\eta}_{p_k},
    \label{eqn:pos_meas}
\end{align}
where $\tilde{\mathbf{p}}_k \in \mathbb{R}^d$ is a measurement of the true position $\mathbf{x}_k \in \mathbb{R}^d$ at time $k$ and $\boldsymbol{\eta}_{p_k} \sim \mathcal{N}(0, \mathbf{R}_{p_k})$ is AWGN of covariance $\mathbf{R}_{p_k}$. The measurement Jacobian in this case is $\mathbf{H}_{p_k} = \mathbf{I}$.
%
%An example of a widely used position-based sensor is the Global Positioning System (GPS).

\subsubsection{Range measurement model} \label{sec:range_meas_model}
In range-based localization, the measurement model consists of the distance measured between a tag and an anchor:
\begin{align}
    r_{a_k} = \| \mathbf{p}_a - \mathbf{x}_k \|_2 + \eta_{r_k},
    \label{eqn:rng_meas}
\end{align}
where $r_{a_k} \in \mathbb{R}$ is the measured distance to anchor $a$ with position $\mathbf{p}_a \in \mathbb{R}^d$, $\| \cdot \|_2$ is the $L_2$ norm, and~$\eta_{r_k} \sim \mathcal{N}(0, \sigma^2_{r_k})$ is AWGN of variance $\sigma^2_{r_k}$. Commonly used technologies for range-based positioning include WiFi, UWB, sonar, radar, and lidar. The corresponding measurement Jacobian is
\begin{align}
    \mathbf{H}_{r_k} = - \frac{(\mathbf{p}_a - \mathbf{x}_k)^T}{\| \mathbf{p}_a - \mathbf{x}_k \|_2}.
    \label{eqn:rng_meas_jac}
\end{align}

We now have the necessary components to apply our optimization problem~\eqref{eqn:param_opt_problem} to trajectory estimation for determining required parameters.

\subsection{Parameter estimation}

\subsubsection{Sensor query schedule estimation} \label{sec:sensor_query_schedule}
As mentioned in Section~\ref{sec:prob_statement}, we want to identify a sensor query schedule that minimizes the number of queries. For a given trajectory duration, this is equivalent to maximizing the duration between subsequent measurement times (see Figure~\ref{fig:method_overview}). If we consider $t_k$ and $t_{k+1}$ to be aligned with successive measurements and as such with consecutive sensor queries, then $\Delta t_{k}$ from~\eqref{eqn:wnov_gp_prior_mean} corresponds to the time duration between subsequent sensor queries. Substituting expressions for the process model Jacobian and the covariance function from~\eqref{eqn:wnov_jac_cov} in~\eqref{eqn:inf_err_acc_mat_const}, we get the following optimization problem
\begin{equation}
    \hspace*{-1em}
    \begin{split}
        \underset{\Delta t_{k} \in \mathbb{R}_{++}}{\max} \quad &  \Delta t_{k} \\
        \mathrm{s.t.} \quad &  \boldsymbol{S}_k(\Delta t_k) \succcurlyeq 0, \quad k=0, 1,2,...
    \label{eqn:meas_freq_opt_dt}
    \end{split}
\end{equation}
where
\begin{align*}
    \scalebox{0.89}{$ \boldsymbol{S}_k(\Delta t_k) = 
    \begingroup
    \setlength\arraycolsep{2pt}
    \begin{bmatrix}
        \Jpred_{\mathbf{x}_k} +\mathbf{Q}_k^{-1} + \mathbb{E} \left[ \mathbf{H}_k^T \mathbf{R}_k^{-1} \mathbf{H}_k \right]  & \mathbf{Q}_k^{-1} \\
       \mathbf{Q}_k^{-1} & \mathbf{Q}_k^{-1} - k_a^{-2} \mathbf{I}
        \end{bmatrix}
    \endgroup \succcurlyeq 0
    $}
\end{align*}
 with $\mathbf{Q}_k = \boldsymbol{Q} \Delta t_k$. According to the principles of disciplined convex programming~\cite{Grant2006},~\eqref{eqn:meas_freq_opt_dt} is not convex in $\Delta t_k$, however, it is convex in $\Delta t_k^{-1}$. We can reformulate the above problem as a convex problem by setting $m_k = \Delta t_k^{-1}$:
\begin{equation}
    \begin{split}
        \underset{m_k \in \mathbb{R}_{++}}{\min} & \quad  m_{k} \\
        \mathrm{s.t.} & \quad \boldsymbol{S}_{k}(m_k) \succcurlyeq 0, \quad k=0, 1,2,...
    \label{eqn:meas_freq_opt_dti}
    \end{split}
\end{equation}
where
\begin{align*}
    \scalebox{0.89}{$
    \boldsymbol{S}_{k}(m_k) = 
    \begingroup
    \setlength\arraycolsep{2pt}
    \begin{bmatrix}
        \Jpred_{\mathbf{x}_k} + \mathbf{Q}_{m_k}^{-1} + \mathbb{E} \left[ \mathbf{H}_k^T \mathbf{R}_k^{-1} \mathbf{H}_k \right]  &  \mathbf{Q}_{m_k}^{-1} \\
        \mathbf{Q}_{m_k}^{-1} &  \mathbf{Q}_{m_k}^{-1} - k_a^{-2} \mathbf{I}
        \end{bmatrix},
    \endgroup
    $}
\end{align*}
with $\mathbf{Q}_{m_k} = \boldsymbol{Q} \, m_k^{-1}$. For the WNOV motion model~\eqref{eqn:wnov_proc_model}, the inequality constraint in~\eqref{eqn:meas_freq_opt_dti} is a linear matrix inequality (LMI) at each time step and consequently~\eqref{eqn:meas_freq_opt_dti} is a semidefinite program (SDP).
We can initialize the recursion for~\eqref{eqn:meas_freq_opt_dti} by \textit{(i)} setting $\Jpred_{\mathbf{x}_{0}} = \mathbb{E} \left[- \frac{\partial^2 \log p(\mathbf{x}_{0})}{\partial \mathbf{x}_{0} \partial \mathbf{x}_{0}^T } \right]$ if the distribution over the initial state, $p(\mathbf{x}_0)$, is known, or \textit{(ii)} setting $\Jpred_{\mathbf{x}_{0}} = k_a^2 \mathbf{I}$ if the initial state is known.  Once $\Delta t_k$ is computed, $\Jpred_{\mathbf{x}_{k+1}}$ can be computed using~\eqref{eqn:recr_pcrb_eqn} for the next iteration.

For robots operating in uniform conditions, we can obtain nominal values for Jacobians. Assuming the initial conditions are known, we can solve for a constant sensor query rate:
\begin{equation}
    \begin{split}
        \underset{m_s \in \mathbb{R}_{++}}{\min} & \quad  m_s \\
        \mathrm{s.t.} & \quad \boldsymbol{S}({m_s}) \succcurlyeq 0,
    \label{eqn:meas_freq_opt_dti_const}
    \end{split}
\end{equation}
where
\begin{align*}
    \scalebox{0.89}{$
    \boldsymbol{S}(m_s) = 
    \begingroup
    \setlength\arraycolsep{2pt}
    \begin{bmatrix}
        \Jpred_{\mathbf{x}_k} + \mathbf{Q}_{m_s}^{-1} + \mathbb{E} \left[ \mathbf{H}_k^T \mathbf{R}_k^{-1} \mathbf{H}_k \right]  &  \mathbf{Q}_{m_s}^{-1} \\
        \mathbf{Q}_{m_s}^{-1} &  \mathbf{Q}_{m_s}^{-1} - k_a^{-2} \mathbf{I}
        \end{bmatrix},
    \endgroup
    $}
\end{align*}
with $\mathbf{Q}_{m_s} = \boldsymbol{Q} \, m_s^{-1}$ and $\Jpred_{\mathbf{x}_k} = k_a^{-2} \mathbf{I}$, since we are solving for a constant sensor query rate. Nominal value for a measurement Jacobian depends on the sensing modality. For instance, for a position sensor, we have $\mathbf{H}_k = \mathbf{H} = \mathbf{I}$. 

\subsubsection{Sensor covariance estimation} \label{sec:sensor_cov_est}
In this section, we want to identify a sensor covariance matrix to achieve a desired accuracy for a given sensor query schedule. We use $\mathbf{R}$ to refer to both $\mathbf{R}_p$ for position sensor covariance and $\sigma_r^2$ for range sensor covariance. Similar to the previous case,~\eqref{eqn:inf_err_acc_mat_const} is not convex in $\mathbf{R}_k$ but is convex in $\mathbf{R}_k^{-1}$. The optimization problem~\eqref{eqn:param_opt_problem} in this case is 
\begin{equation}
    \begin{split}
        \underset{\mathbf{R}_{k}^{-1} \in \mathbb{S}^d_{++}}{\min} & \quad  \mathrm{tr}(\mathbf{R}_k^{-1})  \\
        \mathrm{s.t.} & \quad \boldsymbol{S}_k(\mathbf{R}_k^{-1}) \succcurlyeq 0, \quad k=0, 1,2,...
    \label{eqn:meas_cov_opt}
    \end{split}
\end{equation}
where $\mathrm{tr}(\mathbf{A})$ is the trace of matrix $\mathbf{A}$ and
\begin{align*}
    \scalebox{0.89}{$
    \boldsymbol{S}_k(\mathbf{R}_k^{-1}) = 
    \begingroup
    \setlength\arraycolsep{2pt}
    \begin{bmatrix}
        \Jpred_{\mathbf{x}_k} +\mathbf{Q}_k^{-1} + \mathbb{E} \left[ \mathbf{H}_k^T \mathbf{R}_k^{-1} \mathbf{H}_k \right]  & \mathbf{Q}_k^{-1} \\
       \mathbf{Q}_k^{-1} & \mathbf{Q}_k^{-1} - k_a^{-2} \mathbf{I}
        \end{bmatrix}.
    \endgroup
    $}
\end{align*}
We chose to minimize the trace of the inverse covariance matrix, which corresponds to the $A$-optimal design criteria~\cite[Section 7.5.2]{boydConvexOptimization2004}. However, other design criteria, such as minimizing the determinant ($D$-optimal) or minimizing the norm ($E$-optimal) of $\mathbf{R}^{-1}_k$ can be used while retaining convexity.
In~\eqref{eqn:meas_cov_opt}, we solve for a different measurement covariance at each time instant. Alternatively, we can optimize for a constant measurement covariance for uniform trajectories. 

% \begin{remark}
% Parameters estimated using methods outlined in Sections~\ref{sec:sensor_query_schedule} and~\ref{sec:sensor_cov_est} ensure that the estimation error is within the required bounds in expectation. For instance, results from a single trial could have higher estimation error, but in expectation over the state across multiple trials, the estimation error is within the bounds. 
% \end{remark}

\begin{remark}
    When using sensing modalities with lower dimensionality compared to the state, multiple measurements are needed if $\Jpred_k \preccurlyeq k_a^{-2} \mathbf{I}$. This is expected since a single lower-dimensional measurement cannot constrain the full state. For instance, in 2D trajectory estimation with range sensors, we require at least two anchors to calculate the Jacobian $\mathbf{H}_k$.
\end{remark}

% \subsection{Strong duality}
% The optimization problems~\eqref{eqn:meas_freq_opt_dti} and~\eqref{eqn:meas_cov_opt} are convex semidefinite programs (SDPs) and can be solved in polynomial time to optimality using off-the-shelf solvers. Additionally, if the problems are feasible, \textit{strong duality}~\cite[Section 5.9]{boydConvexOptimization2004} holds if there
% \begin{align*}
%     \mathbf{S}_k(\theta) \succ 0, \quad \theta \in \rm{relint}(\mathcal{D}), \forall \, k,
% \end{align*}
% %
% where $\mathrm{relint(\cdot{})}$ is the relative interior~\cite[Section 2.1.3]{boydConvexOptimization2004} of domain $\mathcal{D}$, with $\mathcal{D} = \mathbb{R}_{++}$ for~\eqref{eqn:meas_freq_opt_dti} and $\mathcal{D} = \mathbb{S}_{++}^d$ for~\eqref{eqn:meas_cov_opt}.  
%
\section{Experiments} \label{sec:experiments}
In this section, we apply our proposed methods to trajectory estimation in simulation and real experiments. We use the CVXPY~\cite{cvxpy} package with MOSEK~\cite{mosek} solver to optimize SDPs for identifying parameters and use Gaussian quadrature to calculate the necessary expectations. We estimate the trajectory of the robot using nonlinear batch maximum a posterior (MAP) inference. We evaluate the performance of MAP inference using root-mean-squared error (RMSE) metric: $\| \mathbf{x}_{\mathrm{gt}}(t) - \mathbf{x}_{\mathrm{est}}(t)\|_2$, where $\mathbf{x}_{\mathrm{gt}}(t)$ and $\mathbf{x}_{\mathrm{est}}(t)$ are the true and the estimated states, respectively. 
\subsection{Simulation} \label{sec:simulation}

\begin{figure}[t]
	\centering
	\includegraphics[scale=0.8]{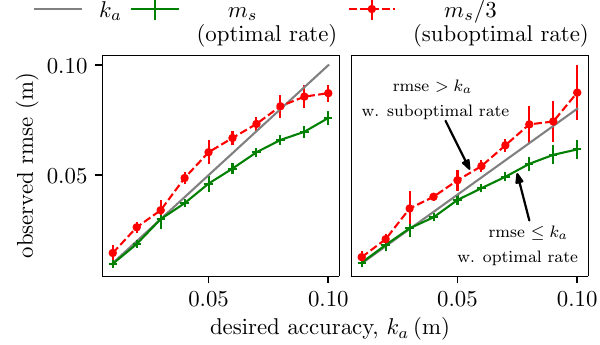}
	\vspace*{-1em}
	\caption{Trajectory estimation results from simulation using position sensors (left) and range sensors (right) with different sensor query rates. Noise values for the position and the range sensors are $\sigma_p = 0.08\,\si{m}$ and $\sigma_r = 0.08\,\si{m}$, respectively. The desired RMSE for a given accuracy value, $k_a$, is indicated by the black line in each case. RMSE values obtained using the proposed optimal sensor query rate, $m_s$ (optimal rate), are equal to or lower than the desired accuracy $k_a$ (black line), whereas the lower sensor query rate, $m_s/3$ (suboptimal rate), results in higher RMSE, indicating the validity of the proposed sensor query rate.}
	\label{fig:rmse_v_freq_sim}
	\vspace*{-1.em}
\end{figure}

\subsubsection{Sensor query schedule estimation} \label{sec:sensor_query_simulation} 
The objective of this simulation is to show that for given sensor covariances, our proposed sensor query schedule achieves the required estimation accuracy. Our approach in simulation is as follows. We first sample random initial positions and velocities between $[-4, 4]\,\si{m}$ and $[-1,1]\,\si{m/s}$, respectively. The initial positions and velocities are used to generate a GP prior by simulating the system dynamics using~\eqref{eqn:wnov_gp_prior_mean}. A ground-truth trajectory is generated by sampling uniformly from the GP prior. The process noise and sensor covariances are selected to reflect real systems and are constant across different experiments: $\boldsymbol{Q} = 0.001 \, \mathbf{I}, \mathbf{R}_p = 0.0064 \, \mathbf{I}$, and $\sigma_r^2 = 0.0064$. For each generated ground-truth trajectory, we calculate a constant sensor query schedule, $m_s$, using~\eqref{eqn:meas_freq_opt_dti_const} for each $k_a \in [0.01, 0.1]\,\si{m}$. In this case, the SDP state size is one (a single scalar) and the average time taken to solve for a constant sensor query rate in~\eqref{eqn:meas_freq_opt_dti_const} is 0.017\,\si{s}. Next, we simulate measurements according to the estimated sensor query schedule. These measurements are subsequently used in the MAP inference to estimate the robot trajectory. The estimated trajectory is compared with the ground trajectory to calculate the associated RMSE.

Average RMSE from 10 trials for each value of $k_a$ for position and range sensors are shown in Figure~\ref{fig:rmse_v_freq_sim}. Error values for a query rate lower than the estimated rate, $m_s/3$, are also shown. In each case, the proposed sensor schedule achieves equal or lower error than the required accuracy and with query rate lower than the identified value, the estimation error is higher than the required accuracy in most cases. However, for $k_a > \sigma_r$ the estimated query schedule is conservative in some cases, indicating that a sensor query schedule with lower query rate could also be viable. 
\subsubsection{Sensor covariance estimation}
In this section, our objective is to show that for a given sensor schedule, our proposed method estimates appropriate sensor covariances to achieve a certain estimation accuracy.

\begin{figure}[t]
	\centering
	\includegraphics[scale=0.8]{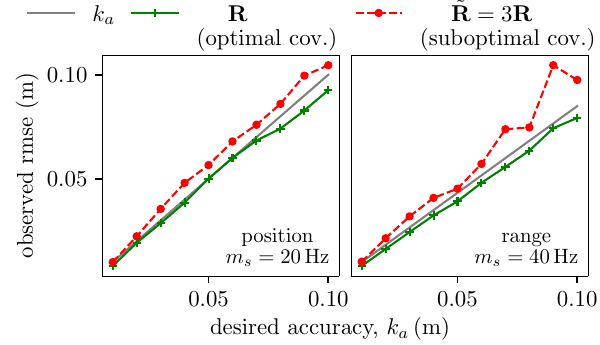}
	\vspace*{-0.8em}
	\caption{Trajectory estimation results from simulation using position sensors (left) and range sensors (right) of proposed covariance, $\mathbf{R}$ (optimal cov.), and higher covariance, $\tilde{\mathbf{R}} = 3\,\mathbf{R}$ (suboptimal cov.). The desired RMSE for a given accuracy value, $k_a$, is indicated by the black line in each case. Using sensors with the proposed covariances, the trajectory estimation RMSE is lower than the desired accuracy $k_a$ in each case. However, using sensors with higher covariances results in RMSE higher than the required accuracy, indicating the validity of our approach.}
	\label{fig:rmse_v_cov_sim}
	\vspace*{-1.5em}
\end{figure}

The setup for this case is similar to the previous case, but now we assume that the sensor covariances are unknown and a sensor query schedule is given. Specifically, we set $m_s = 20\,\si{Hz}$ for positions sensors and $m_s = 40\,\si{Hz}$ for range sensors and for each $k_a \in [0.01, 0.1]\,\si{m}$ we compute the corresponding sensor covariance, $\mathbf{R}$, using~\eqref{eqn:meas_cov_opt}. For 2D trajectory estimation, the state consists of a $2\times 2$ matrix and the average time taken to solve~\eqref{eqn:meas_cov_opt} at each time step is 0.017\,\si{s}. The estimated sensor covariances and the provided sensor schedule are used to simulate measurements that are then used for MAP inference. Results from 10 trials for the different sensing modalities are shown in Figure~\ref{fig:rmse_v_cov_sim}. As before, the identified sensor covariances achieve RMSE lower than the required accuracy. For validation, we also computed RMSE with sensor covariances higher than the one proposed by our methods: $\tilde{\mathbf{R}} = 3\mathbf{R}$. With higher sensor covariance, the estimation error is larger than the required accuracy in most cases, indicating the efficacy of our proposed method.

\subsection{Real experiments} \label{sec:real_experiments}

In real experiments, we apply our proposed method to estimate the trajectory of an omnidirectional mobile robot (see Figure~\ref{fig:method_overview}). The test space is equipped with a motion capture system for ground-truth pose and eight UWB anchors at the corners of the arena. The data rate of the ground-truth pose is $100\,\si{Hz}$, and that of range data is $50\,\si{Hz}$. We performed multiple experiments where the mobile robot was commanded along predefined trajectories, and the sensor data was recorded on the onboard computer for offline evaluation.

\subsubsection{Sensor query schedule estimation}
\begin{figure}[t]
	\includegraphics[scale=0.8]{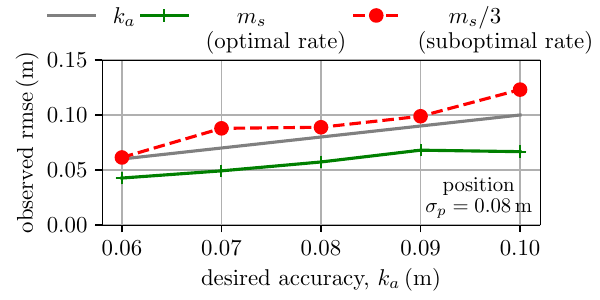}
	\vspace*{-1em}
	\caption{Trajectory estimation RMSE from real experiments with position sensors sampled at the proposed sensor query rate, $m_s$ (optimal rate), and a lower sensor query rate, $m_s/3$ (suboptimal rate). The desired RMSE for different accuracy values, $k_a$, is shown by the gray line. In all cases, the estimated sensor query rate yields RMSE equal to or lower than the desired accuracy $k_a$, whereas the lower sensor query rate results in higher RMSE.}
	\label{fig:pos_rmse_v_acc_real}
	\vspace*{-1.5em}
\end{figure}

We use the commanded trajectory and sensor covariances to calculate a constant sensor query schedule. To simulate noisy position measurements, we corrupt the ground-truth position with Gaussian noise of variance $\mathbf{R}_p = 0.0064\, \mathbf{I}$. The covariance of UWB-based range measurements is calculated using ground truth as $\sigma_r^2 = 0.0049$. To emulate different sensor query rates, we (sub)sampled the recorded data. Average trajectory estimation RMSE from three experiments for $k_a \in [0.01, 0.1]\,\si{m}$ with position measurements and range measurements are shown in Figure~\ref{fig:pos_rmse_v_acc_real} and in Table~\ref{tab:rng_acc_v_rmse_real}, respectively. Estimated trajectories using range measurements from one such experiment are shown in Figure~\ref{fig:method_overview}.

The results highlight the main benefits of our approach. Firstly, for feasible accuracy values, the estimated query rate achieves the desired performance, and a lower query rate, $m_s/3$, results in degraded performance, proving the validity of our approach. Secondly, our proposed method can be used to ascertain if specific accuracy is infeasible given the system parameters. For instance, with the process and sensor covariances in our real setup, the optimization problem for calculating the sensor query schedule for $k_a = 0.001\,\si{m}$ is infeasible (with a certificate of infeasibility from the numerical solver), providing feasible performance limits. Even with a feasible configuration, our approach can be used to identify operational limits. For example the recommended query rate for \textit{(i)} $k_a = 0.03\,\si{m}$ with range sensors is $152\,\si{Hz}$ and \textit{(ii)} $k_a = 0.02\,\si{m}$ with position sensors is $340\,\si{Hz}$ that are beyond the capabilities of the sensors used in our setup.
%As in simulation, we notice that for $k_a > \sigma_r$ the estimated query schedule is conservative for this particular setup. }

% \begin{figure}[t]
% 	\centering
% 	\includegraphics[scale=0.8]{figs/rng_rmse_v_acc_real.pdf}
% 	\vspace*{-1em}
% 	\caption{\lipsum[1]}
% \end{figure}

\begin{table}[b]
	\vspace*{-1em}
	\centering
	\setlength\extrarowheight{2.0pt}
	\setlength{\tabcolsep}{3pt}
	\caption{Trajectory estimation RMSE using range sensors from real experiments. The proposed sensor query rate, $m_s$, achieves RMSE values lower than $k_a$ as desired, whereas the sensor query rate, $m_s/3$ results in RMSE values larger than $k_a$.}
	\vspace*{-0.5em}
	\begin{tabular}{ | c | c | c | c | c |}
		\hline
		\multirow{2}{*}{Query} & \multicolumn{4}{c|}{RMSE\,($\si{m}$)} \\
		\cline{2-5}
		frequency & $k_a = 0.07\,\si{m}$ & $k_a = 0.08\,\si{m}$ & $k_a = 0.09\,\si{m}$ & $k_a = 0.1\,\si{m}$  \\
		\hline
		$m_s$ & \textbf{0.055} & \textbf{0.054} & \textbf{0.067} & \textbf{0.092} \\
		\hline
		${m_s}/{3}$& {0.072} & {0.155} & {0.171}  & {0.156}  \\
		\hline
	\end{tabular}
	\label{tab:rng_acc_v_rmse_real}
\end{table}

\subsubsection{Sensor covariance estimation}
As in simulation, the objective in this section is to estimate sensor covariances to achieve a certain estimation accuracy. We evaluate sensor covariance estimation for position sensors only, as results with range sensors are similar. The sensor query rate is set to $m_s = 20\,\si{Hz}$ for position sensors. Sensor data at the required query rate is generated by downsampling the recorded data. For each $k_a \in [0.01, 0.1]\,\si{m}$, we compute the required sensor covariance $\mathbf{R}_p$ using~\eqref{eqn:meas_cov_opt}. To emulate noisy position sensors, AWGN with the estimated covariance is added to the down-sampled position data. Results from three trials for each value of $k_a$ for the position sensors are shown in Figure~\ref{tab:pos_cov_rmse_real}. 
For feasible parameter configurations, the identified sensor covariances achieve lower RMSE, and higher sensor covariances, $\tilde{\mathbf{R}}_p = 3\,\mathbf{R}_p$, achieve higher RMSE compared to the desired accuracy as shown in Table~\ref{tab:pos_cov_rmse_real}, indicating the validity of our approach.

\begin{table}[t]
	\centering
	\setlength\extrarowheight{2.0pt}
	\setlength{\tabcolsep}{3pt}
	\caption{Trajectory estimation RMSE with position sensors from real experiments. The proposed sensor covariance $\mathbf{R}_p$ achieves RMSE values lower than $k_a$ as desired, whereas the sensor covariance $\tilde{\mathbf{R}}_p = 3 \mathbf{R}_p$ results in RMSE values larger than $k_a$.}
	\vspace*{-0.5em}
	\begin{tabular}{ | c | c | c | c | c |}
		\hline
		\multirow{2}{*}{Sensor} & \multicolumn{4}{c|}{RMSE\,($\si{m}$)} \\
		\cline{2-5}
		covariance & $k_a = 0.07\,\si{m}$ & $k_a = 0.08\,\si{m}$ & $k_a = 0.09\,\si{m}$ & $k_a = 0.1\,\si{m}$  \\
		\hline
		$\mathbf{R}_p$ & \textbf{0.049} & \textbf{0.064} & \textbf{0.074} & \textbf{0.0942} \\
		\hline
	    $\tilde{\mathbf{R}}_p = 3\mathbf{R}_p$ & 0.073 & 0.107 & 0.122  & 0.204  \\
		\hline
	\end{tabular}
	\label{tab:pos_cov_rmse_real}
	\vspace*{-1em}
\end{table}

\section{Conclusion and Future work}

In this work, we presented a framework to estimate sensor parameters to achieve the desired estimation accuracy. We applied our method to calculate the sensor query schedules and the sensor covariances necessary to attain a desired trajectory estimation accuracy. We validated our approach in simulation and real experiments by showing that a desired trajectory estimation RMSE is achieved using the calculated sensor schedules and covariances. The proposed method also identifies scenarios where a certain RMSE is unachievable given the system and sensor parameters. There are several avenues to explore further, such as the inclusion of diverse motion models, estimation of process covariances, and extension to nonlinear manifolds.

%In this work, we presented an estimator-agnostic method to estimate sensor parameters to achieve the desired estimation accuracy. We demonstrated our method by estimating the sensor query schedules and the sensor covariances necessary to achieve a desired trajectory estimation accuracy with a constant-velocity motion model and position and range measurement models. There are several avenues to explore further such as the inclusion of constant-acceleration motion models, estimation of process covariances, and extension to nonlinear manifolds such as the special Euclidean group.
%
\bibliographystyle{unsrt}
\bibliography{main_reference.bib}

% \newpage
% \input{appendix}

\end{document}